\pdfoutput=1

\documentclass[11pt]{article}

\usepackage[]{emnlp2021}

\usepackage{times}
\usepackage{bm}
\usepackage{latexsym}
\usepackage{amsmath}
\usepackage{multirow}
\usepackage{bbm}
\usepackage{graphicx}
\usepackage{amssymb}
\usepackage{booktabs}
\usepackage{adjustbox}
\usepackage{xspace}
\usepackage{times}
\usepackage{url}

\usepackage{array}
\newcommand{\PreserveBackslash}[1]{\let\temp=\\#1\let\\=\temp}
\newcolumntype{C}[1]{>{\PreserveBackslash\centering}p{#1}}
\newcolumntype{R}[1]{>{\PreserveBackslash\raggedleft}p{#1}}
\newcolumntype{L}[1]{>{\PreserveBackslash\raggedright}p{#1}}

\newcommand{\secref}[2][]{Section#1~\ref{sec:#2}}

\newcommand{\tabref}[2][]{Table#1~\ref{tab:#2}}
\newcommand{\figref}[2][]{Figure#1~\ref{fig:#2}}

\definecolor{LightCyan}{rgb}{0.88,1,1}

\usepackage[T1]{fontenc}
\usepackage[utf8]{inputenc}

\usepackage{microtype}

\title{\textsc{IndoBERTweet}: A Pretrained Language Model for Indonesian Twitter with Effective Domain-Specific Vocabulary Initialization}

\author{Fajri Koto \qquad Jey Han Lau \qquad Timothy Baldwin\\
	School of Computing and Information Systems \\
	The University of Melbourne \\
	\texttt{\small ffajri@student.unimelb.edu.au, jeyhan.lau@gmail.com, 
		tb@ldwin.net} \\
}

\begin{document}
\maketitle
\begin{abstract}
	
We present \textsc{IndoBERTweet}, the first large-scale pretrained model 
for Indonesian Twitter that is trained by extending a
monolingually-trained Indonesian BERT model with additive domain-specific vocabulary. We focus in particular on efficient model adaptation under vocabulary
mismatch, and benchmark different ways of initializing the BERT
embedding layer for new word types. We find
that initializing with the average BERT subword embedding makes  
pretraining five times faster, and is more effective than proposed
methods for vocabulary adaptation in terms of extrinsic evaluation over
seven Twitter-based datasets.\footnote{Code and models can be accessed at \url{https://github.com/indolem/IndoBERTweet}}
\end{abstract}

\section{Introduction}

Transformer-based pretrained language models 
\cite{vaswani2017attention,devlin-etal-2019-bert,liu2019roberta,radford2019language} 
have become the backbone of modern NLP systems, due to their success across 
various languages and tasks. However, obtaining high-quality contextualized 
representations for specific domains/data sources such as biomedical, social 
media, and legal, remains a challenge. 

Previous studies
\cite{alsentzer-etal-2019-publicly,chalkidis-etal-2020-legal,nguyen-etal-2020-bertweet}
have shown that for domain-specific text, pretraining from scratch
outperforms off-the-shelf BERT.  As an alternative approach with lower
cost, \citet{gururangan-etal-2020-dont} demonstrated that domain
adaptive pretraining (i.e.\ pretraining the model on target domain text
before task fine-tuning) is effective, although still not as
good as training from scratch.


The main drawback of domain-adaptive pretraining is that domain-specific
words that are not in the pretrained vocabulary are often tokenized
poorly.  For instance, in \textsc{BioBERT} \cite{lee2019biobert},
\textit{Immunoglobulin} is tokenized into \{\textit{I}, \#\#\textit{mm},
\#\#\textit{uno}, \#\#\textit{g}, \#\#\textit{lo}, \#\#\textit{bul},
\#\#\textit{in}\}, despite being a common term
in biology. To tackle this problem,
\citet{poerner-etal-2020-inexpensive,tai-etal-2020-exbert} proposed
simple methods to domain-extend the BERT vocabulary:
\citet{poerner-etal-2020-inexpensive} initialize new vocabulary using a
learned projection from word2vec \cite{mikolov2013efficient}, while
\citet{tai-etal-2020-exbert} use random initialization with weight
augmentation, substantially increasing the number of model parameters. 

New vocabulary augmentation has been also conducted for
language-adaptive pretraining, mainly based on multilingual BERT
(\textsc{mBERT}). For instance, \citet{chau-etal-2020-parsing} replace
99 ``unused'' WordPiece tokens of \textsc{mBERT} with new common tokens
in the target language, while \citet{wang-etal-2020-extending} extend
\textsc{mBERT} vocabulary with non-overlapping tokens
($|\mathbb{V}_{\textsc{mBERT}}-\mathbb{V}_{new}|$). These two approaches
use random initialization for new WordPiece token embeddings.

In this paper, we focus on the task of learning an Indonesian BERT model
for Twitter, and show that initializing domain-specific vocabulary with
average-pooling of BERT subword embeddings is more efficient than
pretraining from scratch, and more effective than initializing based on
word2vec projections \cite{poerner-etal-2020-inexpensive}.  We use
\textsc{IndoBERT} \cite{koto-etal-2020-indolem}, a monolingual BERT for
Indonesian as the domain-general model to develop a pretrained
domain-specific model \textsc{IndoBERTweet} for Indonesian Twitter.


There are two primary reasons to experiment with Indonesian Twitter.
First, despite being the official language of the 5th most populous
nation, Indonesian is underrepresented in NLP (notwithstanding recent
Indonesian benchmarks and datasets
\cite{wilie-etal-2020-indonlu,koto-etal-2020-liputan6,koto-etal-2020-indolem}).
Second, with a large user base, Twitter is often utilized to support
policymakers, business \cite{fiarni2016sentiment}, or to monitor
elections \cite{suciati2019twitter} or health issues
\cite{prastyo2020tweets}. Note that most previous studies that target
Indonesian Twitter tend to use traditional machine learning models (e.g.\
$n$-gram and recurrent models \cite{fiarni2016sentiment,koto2017inset}).

To summarize our contributions: (1) we release \textsc{IndoBERTweet},
the first large-scale pretrained Indonesian language model for social
media data; and (2) through extensive experimentation, we compare a range of
approaches to domain-specific vocabulary initialization over a
domain-general BERT model, and find that a simple average of subword
embeddings is more effective than previously-proposed methods and
reduces the overhead for domain-adaptive pretraining by 80\%.

\section{\textsc{IndoBERTweet}}

\subsection{Twitter Dataset}
\label{sec:twitter}
We crawl Indonesian tweets over a 1-year period using the official
Twitter API,\footnote{\url{https://developer.twitter.com/}} from
December 2019 to December 2020, with 60 keywords covering 4 main topics:
economy, health, education, and government. We found that the Twitter
language identifier is reasonably accurate for Indonesian, and so use it
to filter out non-Indonesian tweets.  From 100 randomly-sampled tweets,
we found a majority of them (87) to be Indonesian, with a small number
being Malay (12) and Swahili (1).\footnote{Note that Indonesian and
  Malay are very closely related, but also that we implicitly evaluate
  the impact of the language confluence in our experiments over (pure)
  Indonesian datasets.}

After removing redundant tweets (with the same ID), we obtain 26M tweets
with 409M word tokens, two times larger than the training data used to
pretrain \textsc{IndoBERT} \cite{koto-etal-2020-indolem}.  We set aside
230K tweets for development, and extract a vocabulary of 31,984 types
based on WordPiece \cite{wu2016google}. We lower-case all words and
follow the same preprocessing steps as English \textsc{BERTweet}
\cite{nguyen-etal-2020-bertweet}: (1) converting user mentions and URLs
into \texttt{@USER} and \texttt{HTTPURL}, respectively; and (2)
translating emoticons into text using the \texttt{emoji}
package.\footnote{\url{https://pypi.org/project/emoji/}}

\subsection{\textsc{IndoBERTweet} Model}


\textsc{IndoBERTweet} is trained based on a masked language model
objective \cite{devlin-etal-2019-bert} following the same procedure as the \texttt{\footnotesize indobert-base-uncased} (\textsc{IndoBERT}) model.\footnote{\url{https://huggingface.co/indolem/indobert-base-uncased}}
It is a transformer encoder with 12 hidden layers (dimension$=$768), 12
attention heads, and 3 feed-forward hidden layers
(dimension$=$3,072). The only difference is the maximum sequence length,
which we set to 128 tokens based on the average number of words per
document in our Twitter corpus.

In this work, we train 5 \textsc{IndoBERTweet} models. The first model
is pretrained from scratch based on the aforementioned
configuration. The remaining four models are based on domain-adaptive
pretraining with different vocabulary adaptation strategies, as discussed in \secref{domain}.



\begin{table*}[]
	\begin{center}
		\begin{adjustbox}{max width=\linewidth}
			\begin{tabular}{llcrrrcl}
				\toprule
				\textbf{Task} & \textbf{Data} & \textbf{\#labels} & \textbf{\#train} & \textbf{\#dev} & \textbf{\#test} & \textbf{5-Fold} & \textbf{Evaluation} \\
				\midrule
				\multirow{2}{*}{{Sentiment Analysis}} &
				IndoLEM \cite{koto-etal-2020-indolem} & 2 &  3,638 & 399 & 1,011 & Yes & F1$_\text{pos}$\\
				& SmSA  \cite{wilie-etal-2020-indonlu}  & 3 & 11,000 & 1,260 & 500 & No & F1$_\text{macro}$\\
				\midrule
				{{Emotion Classification}} &
				EmoT \cite{wilie-etal-2020-indonlu}  & 5 & 3,521 & 440 & 442 & No & F1$_\text{macro}$ \\
				\midrule
				\multirow{2}{*}{{Hate Speech Detection}} &
				HS1 \cite{alfina2017hate}  & 2 & 499 & 72 & 142 & Yes & F1$_\text{pos}$\\
				& HS2 \cite{ibrohim-budi-2019-multi} & 2 &  9,219 & 2,633 & 1,317 & Yes & F1$_\text{pos}$\\
				\midrule
				\multirow{2}{*}{{Named Entity Recognition}} &
				Formal \cite{munarko2018named}  & 3 & 6,500 & 657 & 1,122 & No & F1$_\text{entity}$\\
				& Informal \cite{munarko2018named} & 3 & 6,500 & 657 & 1,227 & No & F1$_\text{entity}$\\
				\bottomrule
			\end{tabular}
		\end{adjustbox}
	\end{center}
	\caption{Summary of Indonesian Twitter datasets used in our experiments.}
	\label{tab:datastat}
\end{table*}

\subsection{Domain-Adaptive Pretraining with Domain-Specific Vocabulary Initialization}
\label{sec:domain}

We apply domain-adaptive pretraining on the domain-general  
\textsc{IndoBERT} \cite{koto-etal-2020-indolem}, which is trained over 
Indonesian Wikipedia, news articles, and an Indonesian web corpus \cite{medved2019indonesian}. Our 
goal is to fully replace \textsc{IndoBERT}'s vocabulary 
($\mathbb{V}_{\text{IB}}$) of 31,923 types with \textsc{IndoBERTweet}'s 
vocabulary ($\mathbb{V}_{\text{IBT}}$) (31,984 types). In 
\textsc{IndoBERTweet}, there are 14,584 (46\%) new types, and 17,400 
(54\%) WordPiece types which are shared with \textsc{IndoBERT}.\footnote{In the implementation, we set the adaptive vocabulary to be the same size with \textsc{IndoBERT} by discarding some ``[unused-\textit{x}]'' tokens of \textsc{IndoBERTweet}.}

To initialize the domain-specific vocabulary, we use \textsc{IndoBERT}
embeddings for the 17,400 shared types, and explore four initialization
strategies for new word types: (1) random initialization from $U(-1,1)$;
(2) random initialization from $\mathcal{N}(\mu,\sigma)$, where $\mu$
and $\sigma$ are learned from \textsc{IndoBERT} embeddings; (3) linear
projection via \texttt{fastText} embeddings
\cite{poerner-etal-2020-inexpensive}; and (4) averaging
\textsc{IndoBERT} subword embeddings.

For the linear projection strategy (Method 3), we train 300d
\texttt{fastText} embeddings \cite{bojanowski-etal-2017-enriching} over
the tokenized Indonesian Twitter corpus. Following
\citet{poerner-etal-2020-inexpensive}, we use the shared types
($\mathbb{V}_{\text{IB}} \cap \mathbb{V}_{\text{IBT}}$) to train a
linear transformation from \texttt{fastText} embeddings
$\text{E}_{\text{FT}}$ to \textsc{IndoBERT} embeddings
$\text{E}_\text{IB}$ as follows:
\begin{align*}
\underset{\text{\bf W}}{\mathrm{argmin}} \sum_{x\in \mathbb{V}_{\text{IB}} \cap \mathbb{V}_{\text{IBT}}} \| \text{E}_{\text{FT}}(x) \: \text{\bf W} - \text{E}_\text{IB}(x) \|_2^2
\end{align*}
where $\text{\bf W}$ is a $\text{dim}(\text{E}_{\text{FT}})$ $\times$ 
$\text{dim}(\text{E}_{\text{IB}})$ matrix.

To average subword embeddings of $x \in \mathbb{V}_{\text{IBT}}$ (Method 4), we compute:
\begin{align*}
\text{E}_{\text{IBT}}(x) = \frac{1}{|\text{T}_{\text{IB}}(x)|} \sum_{y\in \text{T}_{\text{IB}}(x)} \text{E}_{\text{IB}} (y)
\end{align*}
where $\text{T}_{\text{IB}}(x)$ is the set of WordPiece tokens for word $x$ 
produced by \textsc{IndoBERT}'s tokenizer.

\section{Experimental Setup}

We accumulate gradients over 4 steps to simulate a batch size of
2048. When pretraining from scratch, we train the model for 1M
steps, and use a learning rate of $1\mathrm{e}{-4}$ and the Adam optimizer with a linear scheduler.  All pretraining experiments are done using 4$\times$V100 GPUs (32GB).

For domain-adaptive pretraining (using \textsc{IndoBERT} model), we
consider three benchmarks: (1) domain-adaptive pretraining without
domain-specific vocabulary adaptation
($\mathbb{V}_{\text{IBT}}=\mathbb{V}_{\text{IB}}$) for 200K steps; (2)
applying the new vocabulary adaptation approaches from \secref{domain}
without additional domain-adaptive pretraining; and (3) applying
the new vocabulary adaptation approaches from \secref{domain} with 200K
domain-adaptive pretraining steps.


\textbf{Downstream tasks}. To evaluate the pretrained models, we use 7
Indonesian Twitter datasets, as summarized in \tabref{datastat}.  This
includes sentiment analysis
\cite{koto2017inset,purwarianti2019improving}, emotion classification
\cite{saputri2018emotion}, hate speech detection
\cite{alfina2017hate,ibrohim-budi-2019-multi}, and named entity
recognition \cite{munarko2018named}.
For emotion classification, the classes are \texttt{fear}, 
\texttt{angry}, \texttt{sad}, \texttt{happy}, and \texttt{love}.
Named entity recognition (NER) is based on the \texttt{PERSON}, 
\texttt{ORGANIZATION}, and \texttt{LOCATION} tags.  NER has two test set 
partitions, where the first is formal texts (e.g.\ news snippets on 
Twitter) and the second is informal texts. The train and dev
partitions are a mixture of formal and informal tweets, and shared
across the two test sets.

\textbf{Fine-tuning}. For sentiment, emotion, and hate speech
classification, we add an MLP layer that takes the average pooled output
of \textsc{IndoBERTweet} as input, while for NER we use the first
subword of each word token for tag prediction. We pre-process the tweets
as described in \secref{twitter}, and use a batch size of 30, maximum
token length of 128, learning rate of $5\mathrm{e}{-5}$, Adam optimizer
with epsilon of $1\mathrm{e}{-8}$, and early stopping with patience of
5.  We additionally introduce a canonical split for both hate speech
detection tasks with 5-fold cross validation, following
\citet{koto-etal-2020-indolem}.  In \tabref{datastat}, SmSA, EmoT, and
NER use the original held-out evaluation splits.

\textbf{Baselines}. We use the two \textsc{IndoBERT} models from 
\citet{koto-etal-2020-indolem} and \citet{wilie-etal-2020-indonlu} as 
baselines, in addition to multilingual BERT (\textsc{mBERT}, which 
includes Indonesian) and a monolingual BERT for Malay 
(\textsc{malayBERT}).\footnote{\url{https://huggingface.co/huseinzol05/bert-base-bahasa-cased}} 
Our rationale for including \textsc{malayBERT} is that we are interested in testing 
its performance on Indonesian, given that the two languages are closely
related and we know that the Twitter training data includes some amount
of Malay text.



\section{Experimental Results}
\label{sec:result}

\begin{table*}[]
	\begin{center}
		\begin{adjustbox}{max width=\linewidth}
			\begin{tabular}{lcccccccc}
				\toprule
				\multirow{2}{*}{\textbf{Model}} & \multicolumn{2}{c}{\bf Sentiment} & \bf Emotion & \multicolumn{2}{c}{\bf Hate Speech} & \multicolumn{2}{c}{\bf NER} & \multirow{2}{*}{\bf Average}\\
				\cmidrule{2-8}
				& {\bf IndoLEM} & {\bf SmSA} & {\bf EmoT} & {\bf HS1} & {\bf HS2} & {\bf Formal} & {\bf Informal} & \\
				\midrule
				\textsc{mBERT} & 76.6 & 84.7 & 67.5 & 85.1 & 75.1 & 85.2 & 83.2 & 79.6 \\
				\textsc{malayBERT} & 82.0 & 84.1 & 74.2 & 85.0 & 81.9 & 81.9 & 81.3 & 81.5 \\
				\textsc{IndoBERT} \cite{wilie-etal-2020-indonlu} & 84.1 & 88.7 & 73.3 & 86.8 & 80.4 & 86.3 & 84.3 & 83.4 \\
				\textsc{IndoBERT} \cite{koto-etal-2020-indolem} & 84.1 & 87.9 & 71.0 & 86.4 & 79.3 & 88.0 & 86.9 & 83.4 \\
				\textsc{IndoBERTweet} (1M steps) & 86.2 & 90.4 & 76.0 & \bf 88.8 & \bf 87.5 & 88.1 & 85.4 & 86.1 \\
				\midrule
				\multicolumn{9}{l}{\textsc{IndoBERT} \cite{koto-etal-2020-indolem} \textit{+ 200K steps of domain-adaptive pretraining}} \\	
				\midrule
				Same vocabulary ($\mathbb{V}_{\text{IBT}}=\mathbb{V}_{\text{IB}}$) & 86.4 & \bf 92.7 & 76.8 & 88.7 & 82.2 & 87.9 & 86.9 & 85.9 \\
				
				\midrule
				\multicolumn{9}{l}{\textsc{IndoBERT} \cite{koto-etal-2020-indolem}
                          \textit{ + vocabulary adaptation + 0 steps of domain-adaptive pretraining }} \\
				\midrule
				Uniform distribution & 82.9 & 84.6 & 73.2 & 84.9 & 78.2 & 84.3 & 84.4 & 81.8 \\
				Normal distribution & 83.5 & 86.7 & 71.1 & 85.2 & 77.4 & 85.0 & 86.3 & 82.2 \\
				\texttt{fastText} projection & 84.4 & 83.6 & 72.2 & 85.5 & 80.9 & 85.4 & 85.6 & 82.5 \\
				Average of subwords & 84.2 & 88.1 & 71.6 & 86.2 & 78.3 & 86.4 & \bf 87.4 & 83.2 \\
				\midrule
				\multicolumn{9}{l}{
                          \textsc{IndoBERT} \cite{koto-etal-2020-indolem} \textit{+ vocabulary adaptation + 200K steps of domain-adaptive pretraining}} \\
				\midrule
				
				Uniform distribution & 85.6 & 90.9 & 75.7 & 88.4 & 83.0 & 87.7 & 85.9 & 85.3 \\
				Normal distribution & 87.1 & 92.5 & 75.4 & \bf 88.8 & 82.5 & \bf88.7 & 86.6 & 85.9 \\
				\texttt{fastText} projection & 86.4 & 89.7 & 78.5 & 88.7 & 84.4 & 88.0 & 86.6 & 86.0 \\
				Average of subwords & \bf 86.6 & \bf 92.7 & \bf 79.0 & 88.4 & 84.0 & 87.7 & 86.9 & \bf 86.5 \\
			
				\bottomrule
			\end{tabular}
		\end{adjustbox}
	\end{center}
    \caption{A comparison of pretrained models with different adaptive 
    pretraining strategies for Indonesian tweets (\%).}
	\label{tab:result}
\end{table*}

\tabref{result} shows the full results across the different pretrained
models for the 7 Indonesian Twitter datasets. Note that the first four 
models are pretrained models without domain-adaptive pretraining (i.e.\ 
they are used as purely off-the-shelf models).  In terms of baselines,
\textsc{malayBERT} is a better model for Indonesian than \textsc{mBERT},
consistent with \citet{koto-etal-2020-indolem}, and better again are the
two different \textsc{IndoBERT} models at almost identical
performance.\footnote{Noting that \citet{wilie-etal-2020-indonlu}'s
  version includes 100M words of tweets for pretraining, but
  \citet{koto-etal-2020-indolem}'s version does not.}
\textsc{IndoBERTweet} --- trained from scratch for 1M steps --- results in a
substantial improvement in terms of average performance (almost $+$3\%
absolute), consistent with previous findings that off-the-shelf
domain-general pretrained models are sub-optimal for domain-specific
tasks
\cite{alsentzer-etal-2019-publicly,chalkidis-etal-2020-legal,nguyen-etal-2020-bertweet}.

First, we pretrain \textsc{IndoBERT} \cite{koto-etal-2020-indolem}
\textit{without} vocabulary adaptation for 200K steps, and find
that the results are slightly lower than \textsc{IndoBERTweet}. 
In the next set of experiments, we take \textsc{IndoBERT}
\cite{koto-etal-2020-indolem} and replace the domain-general vocabulary
with the domain-specific vocabulary of \textsc{IndoBERTweet}, without
any pretraining (``0 steps''). Results drop overall relative to the
original model, with the embedding averaging method (``{Average of
  Subwords}'') yielding the smallest overall gap of $-$0.2\% absolute.

Finally, we pretrain \textsc{IndoBERT} \cite{koto-etal-2020-indolem} for
200K steps in the target domain, after performing vocabulary
adaptation. We see a strong improvement for all initialization methods,
with the embedding averaging method once again performing the best, in
fact outperforming the domain-specific \textsc{IndoBERTweet} when
trained for 1M steps from scratch. These findings reveal that we can
adapt an off-the-shelf pretrained model very efficiently (5 times faster
than training from scratch) \textit{with better average performance}.


\section{Discussion}

Given these positive results on Indonesian, we conducted a similar
experiment in a second language, English: we follow \citet{nguyen-etal-2020-bertweet} 
in adapting \textsc{RoBERTa}\footnote{The base version.} for Twitter using
the embedding averaging method to initialize new vocabulary, and
compare ourselves against \textsc{BERTweet}
(trained from scratch on 845M English tweets).

A caveat here is that \textsc{BERTweet} \cite{nguyen-etal-2020-bertweet} and \textsc{RoBERTa} \cite{liu2019roberta} use
different tokenization methods: \texttt{byte-level BPE} vs.\
\texttt{fastBPE} \cite{sennrich-etal-2016-neural}.  Because of this,
rather than replacing \textsc{RoBERTa}'s vocabulary with
\textsc{BERTweet}'s (like our Indonesian experiments), we train
\textsc{RoBERTa}'s BPE tokenizer on English Twitter data (described
below) to create a domain-specific vocabulary. This means that the two
models (\textsc{BERTweet} and domain-adapted \textsc{RoBERTa} with
modified vocabulary) will not be directly comparable.


Following \citet{nguyen-etal-2020-bertweet}, we download 42M tweets from
the Internet
Archive\footnote{\url{https://archive.org/details/twitterstream}} over
the period July 2017 to October 2019 (the first two days of each month),
which we use for domain-adaptive pretraining. Note that this pretraining
data is an order of magnitude smaller than that of \textsc{BERTweet}
(42M vs.\ 845M). We use SpaCy\footnote{\url{https://spacy.io/}} to
filter English tweets, and follow the same preprocessing steps and
downstream tasks as \citet{nguyen-etal-2020-bertweet} (7 tasks in total;
see the Appendix for details). We pretrain \textsc{RoBERTa} for 200K
steps using the embedding averaging method.

\begin{table}[t]
	\begin{center}
		\begin{adjustbox}{max width=\linewidth}
            \begin{tabular}{L{6.5cm}c}
				\toprule
				\bf Model & \bf Average \\
				\midrule
				\textsc{RoBERTa} & 72.9 \\
				\textsc{BERTweet} \cite{nguyen-etal-2020-bertweet} & \bf 76.3 \\
				\midrule
                \textsc{RoBERTA} + \textit{vocabulary adaptation + 200K steps of domain-adaptive pretraining} &
                \multirow{2}{*}{74.1} \\
				\bottomrule
			\end{tabular}
		\end{adjustbox}
    \end{center}
    \caption{English results (\%). The presented performance is averaged over 7
      downstream tasks \cite{nguyen-etal-2020-bertweet}.  Refer to the Appendix for details.}
	\label{tab:result_avg_en}
\end{table}

In \tabref{result_avg_en}, we see that \textsc{BERTweet} outperforms
\textsc{RoBERTa} ($+$3.4\% absolute). With domain-adaptive pretraining using
domain-specific vocabulary, the performance gap narrows to $+$2.2\%, but
are not as impressive as our Indonesian experiments. There are two
reasons for this: (1) our domain-adaptive pretraining data is an order
of magnitude smaller than for \textsc{BERTweet}; and (2) the difference
in tokenization methods between \textsc{BERTweet} and \textsc{RoBERTa}
results in a very different vocabulary.

\begin{figure}[t]
	\centering
	\includegraphics[width=3in]{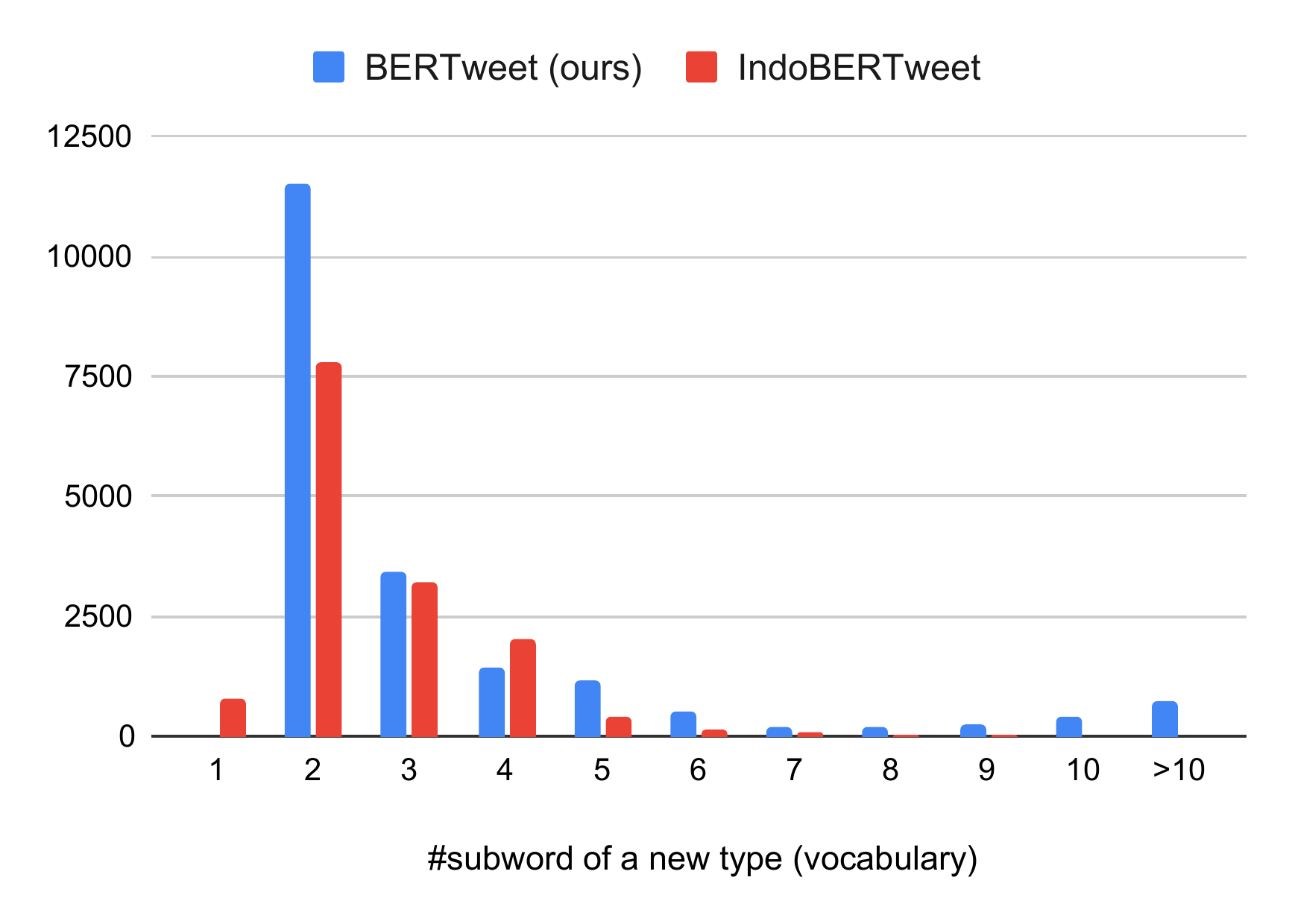}
	\caption{\label{fig:subword} Frequency of \#subword of new types in \textsc{BERTweet} (ours) and \textsc{IndoBERTweet}, tokenized by \textsc{RoBERTa} and \textsc{IndoBERT} tokenizers, respectively. \#subword = 1 means the new type is tokenized as ``[UNK]''.}
\end{figure}

Lastly, we argue that the different  tokenization settings between
\textsc{IndoBERTweet} and \textsc{BERTweet} (ours) may also contribute
to the difference in results. The differences include: (1) uncased vs.\ cased; (2)
WordPiece vs.\ \texttt{fastBPE} tokenizer; and (3) vocabulary size (32K vs.\ 50K)
between both models. In \figref{subword}, we present the frequency
distribution of \#subword of new types in both models after tokenizing by
each general-domain tokenizer. Interestingly, we find that
\textsc{BERTweet} has more new types than \textsc{IndoBERTweet}, with
\#subword after tokenization being more varied (average length of
\#subword of new types are 2.6 and 3.4 for \textsc{IndoBERTweet} and
\textsc{BERTweet}, respectively).



\section{Conclusion}

We present the first large-scale pretrained model for Indonesian
Twitter.  We explored domain-adaptive pretraining with domain-specific
vocabulary adaptation using several strategies, and found that the best
method --- averaging of subword embeddings from the original model ---
achieved the best average performance across 7 tasks, and is five times
faster than the dominant paradigm of pretraining from scratch.

\section*{Acknowledgements}
We are grateful to the anonymous reviewers for their helpful feedback
and suggestions. The first author is supported by the Australia Awards
Scholarship (AAS), funded by the Department of Foreign Affairs and Trade
(DFAT), Australia.  This research was undertaken using the LIEF
HPC-GPGPU Facility hosted at The University of Melbourne. This facility
was established with the assistance of LIEF Grant LE170100200.

\bibliography{anthology,custom}
\bibliographystyle{acl_natbib}

\newpage
\onecolumn
\appendix

\section{Results with English \textsc{BERTweet}}
\label{sec:appendix}

\begin{table*}[ht]
	\begin{center}
		\begin{adjustbox}{max width=\linewidth}
			\begin{tabular}{L{3cm}cccccccc}
				\toprule
				\multirow{2}{*}{\textbf{Model}} & \multicolumn{3}{c}{\bf POS Tagging} & \bf SemEval2017 & \bf SemEval2018 & \multicolumn{2}{c}{\bf NER} & \multirow{2}{*}{\bf Avg.}\\
				\cmidrule{2-8}
				& \bf Ritter11 & \bf ARK & \bf TB-v2 & \bf Sentiment an. & \bf Irony det. & \bf WNUT2016 & \bf WNUT2017 & \\
				\midrule
				\textsc{RoBERTa} & 88.6 & 91.0 & 93.4 & 70.8 & 71.6 & 45.5 & 49.9 & 72.9 \\
				\textsc{BERTweet} & \bf 90.5 & \bf 93.2 & 94.8 & \bf 72.6 & \bf 75.5 & \bf 51.9 & \bf 55.9 & \bf 76.3 \\
				\midrule
				\multicolumn{2}{l}{steps = 0} \\
				\midrule
				\textsc{RoBERTa} w/ \textsc{BERTweet} tokenizer & 87.7 & 90.2 & 92.5 & 65.7 & 66.1 & 39.2 & 39.2 & 68.7 \\
				\textsc{RoBERTa} w/ new tokenizer & 87.4 & 89.6 & 92.8 & 66.5 & 70.7 & 42.5 & 42.4 & 70.3 \\
				\midrule
				\multicolumn{2}{l}{steps = 200K} \\
				\midrule
				\textsc{RoBERTa} w/ new tokenizer & 90.1 & 90.9 & \bf 94.9 & 72.1 & 73.3 & 47.2 & 50.0 & 74.1 \\
				
				\bottomrule
			\end{tabular}
		\end{adjustbox}
	\end{center}
	\caption{English Results (\%) over the test sets. All data,
          metrics, and splits are based off the experiments of
          \citet{nguyen-etal-2020-bertweet}. We re-ran all experiments
          and found slightly lower performance for some models as
          compared to \textsc{BERTweet}. For evaluation, the POS tagging
          datasets
          \cite{ritter-etal-2011-named,gimpel-etal-2011-part,liu-etal-2018-parsing}
          use accuracy, SemEval2017 \cite{rosenthal-etal-2017-semeval}
          uses $\text{Avg}_\text{Rec}$, SemEval2018
          \cite{van-hee-etal-2018-semeval} uses $\text{F1}_\text{pos}$,
          and NER
          \cite{strauss-etal-2016-results,derczynski-etal-2017-results}
          uses $\text{F1}_\text{entity}$.}
	\label{tab:result_en}
\end{table*}

\end{document}